\documentclass[letterpaper, 10 pt, conference]{ieeeconf}  

\IEEEoverridecommandlockouts                              

\usepackage{epsfig}
\usepackage{multirow}
\usepackage{threeparttable}
\usepackage{pifont}
\usepackage{amssymb}
\usepackage{amsmath}
\usepackage{breqn}
\usepackage{url}
\usepackage{makecell}
\usepackage{cite}
\usepackage{textcomp}

\usepackage{hyperref}
\makeatletter
\renewcommand*{\eqref}[1]{%
	\hyperref[{#1}]{\textup{\tagform@{\ref*{#1}}}}%
}
\makeatother

\makeatletter
    \newcommand*\bigcdot{\mathpalette\bigcdot@{.5}}
    \newcommand*\bigcdot@[2]{\mathbin{\vcenter{\hbox{\scalebox{#2}{$\m@th#1\bullet$}}}}}
\makeatother

\usepackage[labelfont={bf},font=small]{caption}

\hypersetup{
	colorlinks=true,
	linkcolor=red,
	filecolor=magenta,      
	urlcolor=blue,
	pdftitle={DeepRelativeFusion: Dense Monocular SLAM using Single-Image Relative Depth Prediction},
}

\title{\LARGE \bf
DeepRelativeFusion: Dense Monocular SLAM using Single-Image Relative Depth Prediction
}

\author{Shing Yan Loo$^{1,2}$, Syamsiah Mashohor$^{2}$, Sai Hong Tang$^{2}$ and Hong Zhang$^{1}$
	\thanks{$^{1}$The authors are with Department of Computing Science, University of Alberta, Canada.}
	\thanks{$^{2}$The authors are with Faculty of Engineering, Universiti Putra Malaysia, Malaysia.}
}

\begin{document}

\onecolumn
\pagestyle{empty} 
\copyright2021 IEEE.  Personal use of this material is permitted.  Permission from IEEE must be obtained for all other uses, in any current or future media, including reprinting/republishing this material for advertising or promotional purposes, creating new collective works, for resale or redistribution to servers or lists, or reuse of any copyrighted component of this work in other works.

Accepted to be published in the Proceedings of the 2021 IEEE/RSJ International Conference on Intelligent Robots and Systems (IROS 2021).
\twocolumn 

\maketitle
\thispagestyle{empty}
\pagestyle{empty}

\maketitle
\thispagestyle{empty}
\pagestyle{empty}

\begin{abstract}

Traditional monocular visual simultaneous localization and mapping (SLAM) algorithms have been extensively studied and proven to reliably recover a sparse structure and camera motion. Nevertheless, the sparse structure is still insufficient for scene interaction, e.g., visual navigation and augmented reality applications. To densify the scene reconstruction, the use of single-image absolute depth prediction from convolutional neural networks (CNNs) for \textit{filling in} the missing structure has been proposed. However, the prediction accuracy tends to not generalize well on scenes that are different from the training datasets.

In this paper, we propose a dense monocular SLAM system, named DeepRelativeFusion, that is capable to recover a globally consistent 3D structure. 
To this end, we use a visual SLAM algorithm to reliably recover the camera poses and semi-dense depth maps of the keyframes, and then use relative depth prediction to densify the semi-dense depth maps and refine the keyframe pose-graph.
To improve the semi-dense depth maps, we propose an adaptive filtering scheme, which is a structure-preserving weighted average smoothing filter that takes into account the pixel intensity and depth of the neighbouring pixels, yielding substantial reconstruction accuracy gain in densification.
To perform densification, we introduce two incremental improvements upon the energy minimization framework proposed by DeepFusion: (1) an improved cost function, and (2) the use of single-image relative depth prediction.
After densification, we update the keyframes with two-view consistent optimized semi-dense and dense depth maps to improve pose-graph optimization, providing a feedback loop to refine the keyframe poses for accurate scene reconstruction.
Our system outperforms the state-of-the-art dense SLAM systems quantitatively in dense reconstruction accuracy by a large margin.

For more information, see the \href{https://www.youtube.com/watch?v=sFuSNKESjzs}{demo video} and \href{https://drive.google.com/file/d/1XlR0wg1uOWHR9B5-TZCgNN3IMEBu-3o1/view?usp=sharing}{supplementary material}.

\end{abstract}

\section{INTRODUCTION}

Recovering dense structure from images can lead to many applications, ranging from augmented reality to self-driving. Visual SLAM uses only cameras to recover structure and motion, which provides cheaper solutions to the SLAM problems in comparison to light detection and ranging (LiDAR).
Traditional monocular visual SLAM algorithms have shown promising sparse~\cite{svo,dso,orbslam,fastorbslam} and semi-dense~\cite{lsdslam} reconstruction accuracy by reliably matching the texture-rich image regions such as corners and edges.
While the sparse structure suffices for localizing the camera, having a dense structure could enable better interaction between a moving robot and the environment, e.g., obstacle avoidance and path planning.

Thanks to the ubiquity of graphics processing units (GPUs), computation of a dense structure from an image sequence in real-time has become possible by aggregating the photometric information in bundles of frames~\cite{dtam}.
In general, the photometric information aggregation seeks to optimize the map by reducing the photometric re-projection errors between bundles of frames, which is a necessary but not sufficient condition to obtain a globally optimized solution.
One inherent limitation is the minimization of photometric re-projection errors in textureless image regions in a bundle of frames as no distinct local minima can be found~\cite{dtam}.
Nevertheless, one common practice in recovering depth information in texture-poor regions is to enforce a \textit{smoothness} constraint~\cite{newcombe2010live,stuhmer2010real}, i.e., the adjacent depth values in the texture-poor image regions change gradually.

\begin{figure}
	\centering
	\includegraphics[scale=1.0]{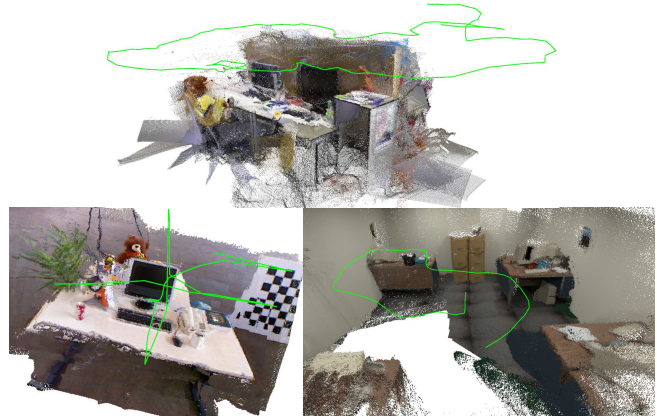} 
	\caption{Qualitative reconstruction of our dense SLAM system on (top) TUM RGB-D~\cite{tumrgbddataset} fr3\_long\_office\_household, (bottom left) TUM RGB-D fr2\_xyz, and (bottom right) ICL-NUIM~\cite{iclnuimdataset} of\_kt2. The green line represents the camera trajectory. Best viewed digitally.}
	\label{fig:main}
\end{figure}

Alternatively, the use of constraints from CNN depth~\cite{cnnslam,deepfusion,deepfactors} and surface normals~\cite{sparse2dense,denseslamnormals} predictions has been proposed to recover the 3D structure in texture-poor image regions.
Both depth and surface normals provide 3D geometry information, the difference being that surface normals contain local surface orientation (i.e., the relative locations between local space points) while a depth value contains the absolute location of a space point.
Therefore, the incorporation of learned 3D geometry into traditional SLAM algorithms have been proposed to solve the monocular dense reconstruction problem.


In this paper, we present a dense SLAM system that augments a monocular SLAM algorithm~\cite{lsdslam} with a dense mapping optimization framework.
The optimization framework exploits the accurate depth and depth gradient information from single-image relative depth prediction as priors to densify the semi-dense structure provided by the SLAM algorithm.
Next, we use the densified structure to refine the keyframe poses, while the optimized poses are combined with the densified structure to produce a globally consistent dense structure (see Figure~\ref{fig:main}).
The experimental results show that our system achieves state-of-the-art dense reconstruction accuracy.
Our main contributions can be summarized as follows:
\begin{itemize}
	\item To the best of our knowledge, we are the first to propose the use of single-image \textit{relative} depth prediction, as opposed to absolute depth prediction, to solve the dense monocular SLAM problem. 
	\item We show, quantitatively and qualitatively, that relative depth maps result in the state-of-the-art reconstruction accuracy.
	\item We introduce a structure-preserving and noise-reducing adaptive filter that improves the accuracy of the semi-dense structure by the monocular SLAM algorithm.
	\item We present a method that makes use of a dense and semi-dense structure to refine the estimated camera motion, to improve its pose estimation.
\end{itemize}

\section{RELATED WORK}

Traditional monocular SLAM algorithms are capable of producing sparse, semi-dense, and dense structures.
Conceptually, sparse refers to the sparsity of the structure as well as the independence of each space point from one another during the structure and motion optimization.
During the optimization, each image point (usually a corner) is being matched across frames and mapped, and collectively, the whole structure and camera motion are being optimized in the form photometric~\cite{dso} or geometric~\cite{orbslam,svo} re-projection error minimization.
On the other hand, instead of processing the sparse points independently, semi-dense and dense methods employ the notion of the neighbourhood \textit{connectedness} of the points.
Dense methods regularize the neighbouring depth pixels using image gradient~\cite{dtam,newcombe2010live,stuhmer2010real}, typically formulated as a \textit{smoothness} term in an energy minimization framework;
whereas the semi-dense method, LSD-SLAM~\cite{lsdslam}, estimates the depth values of the high gradient image regions, thus semi-dense, and regularizes the semi-dense depth map by computing each depth value the weighted average of the neighbouring depth values with the estimated variances as their weight.
In this work, we use LSD-SLAM to reliably recover a semi-dense structure.
Next, we filter the semi-dense structure using contextual information of the local photometric and depth information, which is inspired by the edge-preserving bilateral filtering from Tomasi and Manduchi~\cite{bilateralfiltering}.
Then, we perform densification through regularization of the structure using the filtered semi-dense structure, and depth and depth gradient information from single-image depth prediction.

There are two types of single-image depth predictions: absolute depth prediction and relative depth prediction. 
Absolute depth prediction problem is to train a CNN to predict the metric depth maps from single images~\cite{semidepth,monodepth,qigeonet,vnlnet}. 
Because of the CNN prediction range, the CNN training is commonly limited to one scene type, e.g., indoor or outdoor dataset.
On the other hand, relative depth prediction is concerned with the estimation of the distance of one space point with respect to the others, i.e., their order in depth, rather than the absolute depth.
Early work on relative depth prediction learns from ordinal depth annotations (closer/farther relationship between two points), which contain fairly accurate sparse depth relationships covering a wide range of scene types (e.g., mixing indoor and outdoor scenes in a combined training dataset) ~\cite{zoranordinal,depthinthewild}.
The training results demonstrate accurate ordinal depth prediction quantitatively on different datasets and qualitatively on unconstrained photos taken from the internet, albeit the absence of absolute depth values.
To train on larger and diverse datasets, Lasinger et al. propose to train a relative depth prediction CNN, named MiDaS~\cite{midas}, using a scale- and shift-invariant loss, which handles unknown depth scale and global shift factors in different datasets.
By eliminating the absolute scale and shift, the MiDaS's relative depth prediction is essentially constrained to disparity space, and is akin to having surface normals prediction~\cite{diversedepth} for regularization of neighbouring space points~\cite{sparse2dense,denseslamnormals}, and therefore is particularly suitable for our semi-dense structure densification framework.

Fusions of single-image depth prediction to visual SLAM algorithms have been proposed to solve dense reconstruction problems.
One approach to performing depth fusion from multiple viewpoints is through the accumulation of probabilistic distribution of depth observations from the single-image depth prediction~\cite{cnnslam,rgbdsensing}.
Recently, Czarnowski et al. propose a factor-graph optimization framework named DeepFactors~\cite{deepfactors}, which jointly optimizes the camera motion and the code-based depth maps.
Each depth map is parameterized in an $n$-dimensional code to avoid costly per-pixel depth map optimization.
Another dense SLAM system proposed by Laidlow et al., named DeepFusion~\cite{deepfusion}, uses the depth and depth gradient predictions from a CNN to constrain the optimized depth maps.
Our proposed system is similar to DeepFusion, except for three key differences: (1) we use depth and depth gradient from relative depth prediction as priors in depth map optimization, (2) through extensive experimentation, we have a better cost function for performing densification, and (3) we use the densified depth maps to refine the camera pose.

\section{METHOD}

\begin{figure}
	\centering
	\includegraphics[scale=1.0]{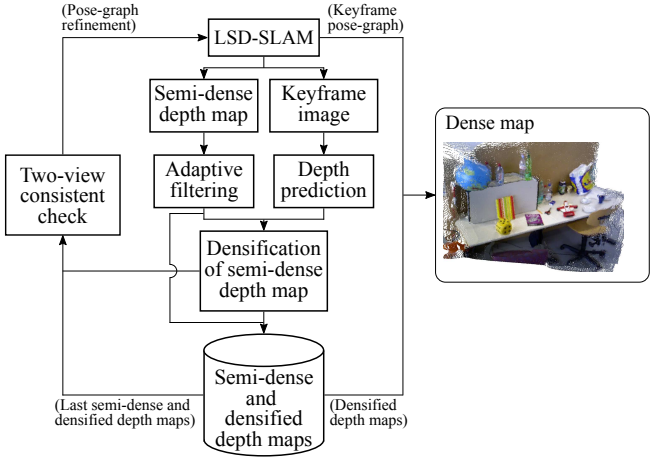}
	\caption{Our dense monocular SLAM pipeline. We augment the LSD-SLAM~\cite{lsdslam} with a depth prediction module, an adaptive filtering module, and a dense mapping module. The optimized depth maps are being to refine the keyframe pose-graph, while the optimized keyframe pose-graph is combined with the densified depth maps to generate a globally consistent 3D structure.  }
	\label{fig:system_overview}
\end{figure}

Our proposed dense SLAM system is shown in Figure~\ref{fig:system_overview}.
The system pipeline contains an optimization framework, which uses the predicted depth maps of the keyframe images (see Section~\ref{subsec:depth_prediction}) and the filtered semi-dense depth maps (see Section~\ref{subsec:adaptive_filtering}) to perform densification (see Section~\ref{subsec:densification}).
The optimized depth maps, in turn, are being used to optimize the keyframe pose-graph maintained by LSD-SLAM (see Section~\ref{subsec:pose_graph_opt}).
To reconstruct the scene, we back-project the densified depth maps from their respective poses obtained from the optimized keyframe pose-graph.

\subsection{Notation}

In LSD-SLAM, the trajectory of the camera poses and the 3D location of the map points are stored in a list of keyframes.
Each keyframe $ \mathcal{K}_i $ contains an image $ I_i :\Omega \rightarrow \mathbb{R} $, 
a semi-dense inverse depth map $ D_{i, \textup{semi-dense}} : \Omega_i \rightarrow \mathbb{R}^+ $,
a semi-dense inverse depth variance map $ V_{i, \textup{semi-dense}} : \Omega_i \rightarrow \mathbb{R}^+ $,
and a camera pose $ S_i \in \textup{Sim}(3)$.
Note that $ \Omega_i \subset \Omega $ is a subset of pixels extracted from the texture-rich image regions for the structure and camera motion estimation, and a Sim(3) camera pose $ S_i $ is defined by:
\begin{equation}
S_i = 
\begin{bmatrix}
s R & t \\
0 & 1
\end{bmatrix},
\end{equation}
where $ R \in SO(3) $ is the rotation matrix, $ t \in \mathbb{R}^3 $ the translation vector and $ s \in \mathbb{R}^+ $ the scaling factor.

\subsection{Depth prediction} \label{subsec:depth_prediction}

For every new keyframe $ \mathcal{K}_i $, we obtain a relative inverse depth map, hereinafter referred to as relative depth map, from MiDaS~\cite{midas} for the densification of the semi-dense depth map.
Because the depth prediction $ D_{i, \textup{CNN}} $ is a relative depth map, the predicted depth map needs to be scale- and shift-corrected before it can be used in the densification step.
The scale- and shift-correction can be performed as follows:
\begin{equation}
D_{i, \textup{CNN}}^\prime = a D_{i, \textup{CNN}} + b,
\end{equation}
where $ a \in \mathbb{R}^+ $ and $ b \in \mathbb{R} $ are the scale and shift parameters, respectively. 
Let $ \vec{d_n} = \begin{pmatrix} d_n & 1 \end{pmatrix}^T $ and $ h^{\textup{opt}} = \begin{pmatrix} a & b \end{pmatrix}^T $, and the parameters $ a $ and $ b $ can be solved in closed-form as follows~\cite{midas}:
\begin{equation} \label{eqn:scaleshift}
h^{\textup{opt}} = \left( \sum_{n \in \Omega_i} \vec{d_n} \vec{d_n}^T \right)^{-1} \left( \sum_{n \in \Omega_i} \vec{d_n} d_n^{\prime} \right),
\end{equation}
where $ d_n \in D_{i, \textup{semi-dense}} $ and  $ d_n^{\prime} \in D_{i, \textup{CNN}} $ are the inverse depth values of the semi-dense depth map and relative depth map, respectively. 

\subsection{Semi-dense structure adaptive filtering} \label{subsec:adaptive_filtering}

Our adaptive filtering is built upon bilateral filtering~\cite{bilateralfiltering}. A bilateral filter is designed to combine the local pixel values according to the geometric closeness $ w_d(.,.) $ and the photometric similarity $ w_s(.,.) $ between the centre pixel $ x $ and a nearby pixel $ x_n $ within a window $ \mathcal{N} $ of an image $ I $, which is defined by:
%
\begin{equation}
\begin{gathered}
    I_{\textup{filtered}}(x) =  \frac{1}{W_{\mathcal{N}}} \sum_{n \in \mathcal{N}} \bigg( I(x_n) \\
    \underbrace{ \exp \Big(
    -\frac{(I(x) - I(x_n))^2}{2\sigma_s^2} \Big)
    }_{=: w_s(x, x_n)}  \underbrace{  \exp \Big(
    - \frac{\left\| x-x_n \right\|^2}{2\sigma_d^2} \Big)
    }_{=: w_d(x, x_n)} \bigg)
\end{gathered}
\end{equation}
%
with
\begin{equation}
W_{\mathcal{N}} = \sum_{n \in \mathcal{N}} w_s(x, x_n) w_d(x, x_n).
\end{equation}

In the context of semi-dense depth map filtering, we introduce two additional weighting schemes, namely, CNN depth consistency $ w_c(.,.) $ and depth uncertainty $ w_u(.) $, to remove the semi-dense depth pixels that have large local variance compared to their corresponding CNN depth as well as large depth uncertainty:
\begin{equation}
\begin{gathered}
w_c(x,x_n) = \exp \left( - \frac{ \Big( \frac{D_{i, \textup{semi-dense}}(x)}{D_{i, \textup{semi-dense}}(x_n)} - \frac{D_{i, \textup{CNN}}^\prime(x)}{D_{i, \textup{CNN}}^\prime(x_n)} \Big)^2 }{2\sigma_c^2} \right) \\
w_u(x_n) = \exp \left( - \frac{\sigma_u V_{i, \textup{semi-dense}}(x_n)}{D_{i, \textup{semi-dense}}(x_n)^4}  \right),
\end{gathered}
\end{equation}
where the \textit{squared ratio difference} in $w_c(.,.)$ computes the scale-invariant error~\cite{eigendepth}, and $ D_{i, \textup{semi-dense}}(x_n) $ in $ w_u(.) $ detects spurious depth pixels. 
$ \sigma_s $, $ \sigma_d $, $ \sigma_c $ and $ \sigma_u $ are the smoothing parameters in their respective spatial kernels. 
Therefore, a filtered semi-dense depth map $ D_{i, \textup{semi-dense}}^\prime $ can be computed as follows:
\begin{align}
\begin{split}
D_{i, \textup{semi-dense}}^\prime(x_n) =  \frac{1}{W_{\mathcal{N}}^\prime} \sum_{n \in \mathcal{N}} \Big( D_{i, \textup{semi-dense}}(x_n) \\ w_s(x, x_n) w_d(x, x_n) w_c(x,x_n) w_u(x_n) \Big)
\end{split}
\end{align}
with
\begin{equation}
W_{\mathcal{N}}^\prime = \sum_{n \in \mathcal{N}} w_s(x, x_n) w_d(x, x_n) w_c(x,x_n) w_u(x_n),
\end{equation}
and, with the updated $ D_{i, \textup{semi-dense}}^\prime $, we re-estimate its semi-dense depth variance map $ V_{i, \textup{semi-dense}}^\prime $ by taking the average of squared deviations within the local window for all the semi-dense depth pixels:
\begin{align}
\begin{split}
V_{i, \textup{semi-dense}}^\prime(x_n) = \frac{\lvert \mathcal{N} \rvert}{n_{\textup{valid}}} \frac{1}{W_{\mathcal{N}}^\prime} \sum_{n \in \mathcal{N}} \Big( W_{\mathcal{N}}^\prime(x_n) \\
\left( D_{i, \textup{semi-dense}}^\prime(x) - D_{i, \textup{semi-dense}}(x_n) \right)^2 \Big),
\end{split}
\end{align}
where $ \lvert \mathcal{N} \rvert $ is the total number of pixels within the window, $ n_{\textup{valid}} $ the number of pixels containing depth values, and $ W_{\mathcal{N}}^\prime(.) $ the weight computed at a nearby pixel.
To remove the noisy depth pixels, we only include filtered depth whose variance is lower than a threshold $ \gamma $.
To ensure similar weighting effect of the semi-dense depth maps in densification, we rescale the semi-dense depth variance $ V_{i, \textup{semi-dense}}^\prime $:
\begin{equation}
V_{i, \textup{semi-dense}}^\prime = \frac{\overline{V_{i, \textup{semi-dense}}}}{\overline{V_{i, \textup{semi-dense}}^\prime}} V_{i, \textup{semi-dense}}^\prime,
\end{equation}
where $ \overline{\cdot} $ is the mean operator.

\subsection{Densification of the semi-dense structure} \label{subsec:densification}

Consider the densification of $ D_{i, \textup{semi-dense}}^\prime $ of $ \mathcal{K}_i $ using $ D_{i, \textup{CNN}}^\prime $ as initial values: the estimated inverse dense depth map $ D_{i, \textup{opt}} $ can be obtained through the minimization of the cost function given by:
\begin{equation}
E_{\textup{total}} = E_{\textup{CNN\_grad}} + \lambda E_{\textup{semi-dense}}.
\end{equation}

The first term, CNN depth gradient regularization $E_{\textup{CNN\_grad}}$, enforces depth gradient consistency between $D_{i, \textup{CNN}}$ and $D_{i, \textup{opt}}$:
\begin{equation} \label{eqn:cnn_grad}
E_{\textup{CNN\_grad}} = \frac{1}{\lvert \Omega \rvert} \sum_{n \in \Omega} \frac{\left(E_{\textup{CNN\_grad}, x}(n) \right)^2 + \left( E_{\textup{CNN\_grad}, y}(n) \right)^2}{ \left( 1 / {D_{i, \textup{CNN}}^\prime(n)} \right)^2} ,
\end{equation}
with
\begin{equation}
\begin{gathered}
E_{\textup{CNN\_grad}, x} = \partial_x \ln D_{i, \textup{opt}} - \partial_x \ln D_{i, \textup{CNN}}^\prime\\
E_{\textup{CNN\_grad}, y} = \partial_y \ln D_{i, \textup{opt}} - \partial_y \ln D_{i, \textup{CNN}}^\prime,
\end{gathered}
\end{equation}
where $ \lvert \Omega \rvert $ is the cardinality of $ \Omega $, and $ \partial $ the gradient operator.
This error term is similar to the \textit{scale-invariant mean squared error in log space} used in~\cite{eigendepth}.
The denominator $\left( 1 / {D_{i, \textup{CNN}}^\prime} \right)^2$ in Equation~\eqref{eqn:cnn_grad} simulates the variance of the depth prediction, which provides stronger depth gradient regularization to closer objects than farther objects.

The second term, semi-dense depth consistency $ E_{\textup{semi-dense}} $, minimizes the difference between the optimized depth map and the semi-dense depth map from LSD-SLAM (similar to~\cite{deepfusion}):
\begin{equation}
E_{\textup{semi-dense}} = \frac{1}{\lvert \Omega_i \rvert} \sum_{n \in \Omega_i} \rho \left( \frac{\left( D_{i, \textup{opt}}(n) - D_{i, \textup{semi-dense}}^\prime(n) \right)^2}{V_{i, \textup{semi-dense}}^\prime(n)} \right),
\end{equation}
where $ \lvert \Omega_i \rvert $ is the cardinality of $ \Omega_i $. 
We add the generalized Charbonnier penalty function~\cite{robustloss}, $ \rho (.) $, to improve reconstruction accuracy.



\subsection{Pose-graph refinement} \label{subsec:pose_graph_opt}

To incorporate the optimized semi-dense and dense structure into improving the keyframe poses while at the same time minimizing the influence of erroneous regions within the structure, we introduce a two-view consistency check step to filter the inconsistent depth regions between the current keyframe $ \mathcal{K}_i $ and the last keyframe $ \mathcal{K}_{i-1} $.
To check for structural consistency, we project the last keyframe semi-dense $ D_{i-1, \textup{semi-dense}}^\prime $ and densified $ D_{i-1, \textup{opt}} $ depth maps to the current keyframe's viewpoint:
\begin{equation} 
\begin{gathered}
\hat{\dot{x}} = K S_{i-1 \rightarrow i} D_{i-1, \bigcdot}(x) K^{-1}  \dot{x} \qquad \textup{with} \quad  \{ x \vert D_{i-1, \bigcdot}(x) > 0 \}\\
\hat{D_{i, \bigcdot}}(\hat{x}) = \left[\hat{\dot{x}} \right]_3,
\end{gathered}
\end{equation}
where $ D_{i-1, \bigcdot} $ is a placeholder for $ D_{i-1, \textup{semi-dense}}^\prime $ and $ D_{i-1, \textup{opt}} $ and $ \hat{D_{i, \bigcdot}}(\hat{x}) $ the warped $ D_{i-1, \textup{semi-dense}}^\prime $ and $ D_{i-1, \textup{opt}} $ in $ \mathcal{K}_i $'s viewpoint.
To retain the semi-dense structure in LSD-SLAM, the semi-dense depth regions in $ D_{i-1, \textup{semi-dense}}^\prime $ has been excluded when warping $ D_{i-1, \textup{opt}}^\prime $.
$ \dot{x} $ is $ x $ in homogeneous coordinates, and $ K $ is the camera intrinsics.
The two-view consistent depth map $ D_{i, \textup{c}} $ is given by:
\begin{equation}
D_{i, \textup{c}}(x) = 
    \begin{cases}
        D_{i, \bigcdot}(x) & \text{if $ \left| D_{i, \bigcdot}(x) - \hat{D_{i, \bigcdot}}(x) \right| < \tau_{e} $}\\
        0 & \text{otherwise}
    \end{cases}.
\end{equation}
Next, we propagate the pose uncertainty $ \Sigma_{\xi, i} \in \mathbb{R}^{7\times7} $ estimated by LSD-SLAM to approximate the uncertainty $ V_{i, opt} $ associated with $ D_{i, \textup{opt}} $:
\begin{equation}
    V_{i, \textup{opt}} \approx J_d \Sigma_{\xi, i} J_d^T
\end{equation}
where $ J_d \in \mathbb{R}^{1\times7} $ is the Jacobian matrix containing the first-order partial derivatives of the camera projection function with respect to the camera pose~\cite{strasdat_thesis}.
As the two-view consistent depth $ D_{i, c} $ contains a mixture of semi-dense depth regions and densified depth regions, the corresponding variance $ V_{i, c} $ is sampled from $ V_{i, \textup{semi-dense}}^\prime $ in the semi-dense depth regions and from $ V_{i, \textup{opt}} $ otherwise.
After obtaining two-view consistent depth $ D_{i, c} $ and depth variance $ V_{i, c} $ maps, we update the Sim(3) constraints in the pose-graph to refine the keyframe poses.

\section{IMPLEMENTATION} \label{sec:implementation}

Our dense SLAM pipeline is implemented using PyTorch~\cite{pytorch} Multiprocessing\footnote{\url{https://pytorch.org/docs/stable/multiprocessing.html}}, which allows for parallel processing of the depth prediction module and the dense mapping module.
To speed up computation, we use Boost.Python\footnote{\url{https://github.com/boostorg/python}} to process loops and deserialize ROS messages\footnote{\url{http://wiki.ros.org/msg}}.

To perform semi-dense structure adaptive filtering, we use a local window size of $ 5 \times 5 $ and the following parameter values: $ \sigma_s = 76.5, \sigma_d = 2, \sigma_c = 0.3, \sigma_u = 2, \gamma = 0.0025$ and $ \beta = 1.1 $.

For the energy minimization, we use PyTorch Autograd~\cite{pytorchautodiff} with Adam optimizer~\cite{adamoptimizer}, where the learning rate is set to 0.05.
To compute the cost function, we set the weighting of the error terms to $ \lambda = 0.003 $, and the generalized Charbonnier function~\cite{robustloss} parameters are set to $ \epsilon = 0.001 $ and $ \alpha = 0.45 $. 
The number of optimization iterations is set to 30. 
The images have been resized to $ 320 \times 240 $ before the depth prediction and densification steps.

For obtaining two-view consistent depth regions, we use an error threshold of $ \tau_{e} = 0.001 $.

In LSD-SLAM, we use the original parameter settings with the exception of setting the \texttt{minUseGrad} parameter to 1 for the following sequences: ICL/office0, ICL/living1, and TUM/seq2 (see Table~\ref{table:reconstruction_results}) and both \texttt{KFUsageWeight} and \texttt{KFDistWeight} parameters to 7.5.
The frame-rate of all image sequences is set to 5 to allow for better synchronization between the camera tracking and the visualization of the dense map;
the increase in frame-rate theoretically should not affect the dense reconstruction performance except for the delayed visualization of the dense map, thanks to the Multiprocessing implementation.

\begin{table*}
	\begin{center}
		\caption{Comparison of overall reconstruction accuracy on the ICL-NUIM dataset~\cite{iclnuimdataset} and the TUM RGB-D dataset~\cite{tumrgbddataset}. (TUM/seq1: fr3\_long\_office\_household, TUM/seq2: fr3\_nostructure\_texture\_near\_withloop, TUM/seq3: fr3\_structure\_texture\_far.)  }
		\label{table:reconstruction_results}
		\begin{tabular}{c | c c c c c c } 
			\Xhline{2\arrayrulewidth}
			& \multicolumn{6}{c }{Percentage of correct depth (\%)} \\
			\cline{2-7}
			Sequence        & CNN-SLAM & DeepFactors*   & DeepFusion    & DeepFusion$ ^\dagger $(MiDaS)*   & Ours (VNLNet)*    & Ours (MiDaS)*     \\ 
			\hline
			ICL/office0     & 19.410   & \textbf{30.17} & 21.090        & 15.934                    & 17.395            & 17.132            \\ 
			ICL/office1     & 29.150   & 20.16          & 37.420        & 57.097                    & \textbf{60.909}   & 58.583            \\  
			ICL/office2     & 37.226   & -              & 30.180        & \textbf{72.602}           & 68.914            & 72.527            \\
			ICL/living0     & 12.840   & 20.44          & 24.223        & 65.395                    & 60.210            & \textbf{65.710}   \\
			ICL/living1     & 13.038   & 20.86          & 14.001        & 75.631                    & 69.980            & \textbf{75.694}   \\ 
			ICL/living2     & 26.560   & -              & 25.235        & 79.994                    & 78.887            & \textbf{80.172}   \\
			TUM/seq1        & 12.477   & 29.33          &  8.069        & \textbf{69.990}           & 64.862            & 66.892            \\
			TUM/seq2        & 24.077   & 16.92          & 14.774        & 52.132                    & 43.607            & \textbf{59.744}   \\
			TUM/seq3        & 27.396   & 51.85          & 27.200        & \textbf{76.433}           & 75.680            & 76.395            \\ 
			\hline
			Average          & 22.464  & 27.10          & 22.466        & 62.801                    & 60.049            & \textbf{63.650}   \\
			\Xhline{2\arrayrulewidth}
		\end{tabular}
		\begin{tablenotes} 
			\item *After aligned with ground truth scale
			\item $^\dagger$Our implementation of DeepFusion
			
		\end{tablenotes}
	\end{center}
\end{table*}

\section{EVALUATION}

In this section, we present experimental results that validate the effectiveness of our proposed method, namely (1) the adaptive filtering to improve the semi-dense depth maps for more accurate densification, (2) the cost function in our optimization framework, (3) the use of relative depth prediction for providing depth and depth gradient priors, and (4) the use of optimized depth maps to improve the keyframe poses.

\subsection{Reconstruction accuracy}
To evaluate our system, we use ICL-NUIM~\cite{iclnuimdataset} and TUM RGB-D~\cite{tumrgbddataset} datasets, which contain ground truth depth maps and trajectories to measure the reconstruction accuracy.
We use the reconstruction accuracy metric proposed in~\cite{cnnslam}, which is defined as the percentage of the depth values with relative errors of less than $10\%$.
Also, we use absolute trajectory error (ATE) to measure the error of the camera trajectories.
Since our system does not produce absolute scale scene reconstruction, and therefore each depth map needs to be scaled using the optimal trajectory scale (calculated with the TUM benchmark script\footnote{\url{https://vision.in.tum.de/data/datasets/rgbd-dataset/tools}}) and its corresponding Sim(3) scale for depth correctness evaluation.

\begin{figure*}
	\centering
	\includegraphics[scale=0.8]{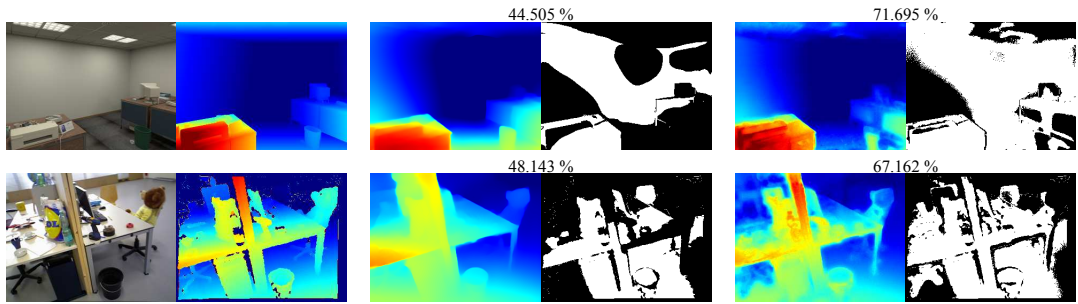}
	\caption{Demonstration of the effectiveness of our optimization framework by comparing the relative depth prediction accuracy from MiDaS before the densification with the densified depth map. 
		(Left column) image and ground truth depth map. (Middle column) scale- and shift-corrected relative depth map and depth correctness mask. (Right column) densified depth map and depth correctness mask. The percentage of correct depth of the depth correctness mask is shown above. }
	\label{fig:optimization_steps}
\end{figure*}

We compare our reconstruction accuracy against the state-of-the-art dense SLAM systems, namely CNN-SLAM~\cite{cnnslam}, DeepFusion~\cite{deepfusion}, and DeepFactors~\cite{deepfactors}.
Table~\ref{table:reconstruction_results} shows a comparison of the reconstruction accuracy:
the first three columns show the reconstruction accuracy of the state-of-the-art systems and the last two columns show a comparison between using VNLNet (an absolute depth prediction CNN) and MiDaS (a relative depth prediction CNN) in our optimization framework (see Section~\ref{subsec:ord_vs_abs_depth}).
Owing to the similarity of the optimization frameworks between our system and DeepFusion, we also include the results for running dense reconstruction with an additional CNN depth consistency error term in the cost function (labelled "$\dagger$" in Table~\ref{table:reconstruction_results})\footnote{DeepFusion is not open source, and therefore the results are based on the implementation of our optimization framework (see Section~\ref{sec:implementation}). 
Our implementation of CNN depth consistency term is similar to that of DeepFusion except we use CNN depth for providing depth uncertainty (similar to Equation~\eqref{eqn:cnn_grad}).}.
Note that the reconstruction accuracy of our method is taken with an average of 5 runs.
Our method outperforms the competitors except for the ICL/office0 sequence, as LSD-SLAM is unable to generate a good semi-dense structure under rotational motion, hence the degraded reconstruction performance in the densification of the semi-dense structure.
The reconstruction results demonstrate the superiority of our system by comparing the last column with all other columns in Table~\ref{table:reconstruction_results}.
Figure~\ref{fig:optimization_steps} shows the use of our optimization framework to obtain more accurate densified depth maps from less accurate predicted relative depth maps.

\begin{table}
	\begin{center}
		\caption{Effect of the error terms on the reconstruction accuracy. (TUM/seq1: fr3\_long\_office\_household, $\circ$: our cost function, $\diamond$: simulated DeepFusion~\cite{deepfusion} cost function, $^\dagger$: not used in DeepFusion.)}
		\label{table:error_terms_ablation}
		\begin{tabular}{c | c c c} 
			\Xhline{2\arrayrulewidth}
			& \multicolumn{3}{c }{Percentage of correct depth (\%)} \\
			\cline{2-4}
			
			Energy term                        & ICL/living2       & ICL/office2       & TUM/seq1              \\
			\hline
			1                                  & 62.620            & 57.563            & 55.031                \\
			\textbf{1(c)}                      & 65.611            & 57.644            & 55.042                \\
			\textbf{1(f)(c)}                   & 71.265            & 61.445            & 60.143                \\
			\hline
			1(c)+2$^\circ$              & 69.967            & 69.905            & 64.650                \\
			\textbf{1(f)(c)+2}$^\circ$  & \textbf{79.788}   & \textbf{71.778}   & 67.319                \\
			\hline
			1(c)+2+3$^\diamond$           & 70.167            & 69.863            & 64.730                \\
			1(f)(c)+2+3$^\diamond$        & 79.742            & 71.692            & \textbf{67.323}       \\
			\Xhline{2\arrayrulewidth}
			\multicolumn{4}{l}{1. SLAM depth consistency}                                                      \\
			\multicolumn{4}{l}{2. CNN depth gradient consistency}                                                    \\
			\multicolumn{4}{l}{3. CNN depth consistency}                                                       \\
			\multicolumn{4}{l}{(c). Generalized Charbonnier function$^\dagger$}                                          \\ 
			\multicolumn{4}{l}{(f). Adaptive semi-dense depth filtering$^\dagger$}                                       \\
			
		\end{tabular}
	\end{center}
\end{table}

\begin{figure}
	\centering
	\includegraphics[scale=1.0]{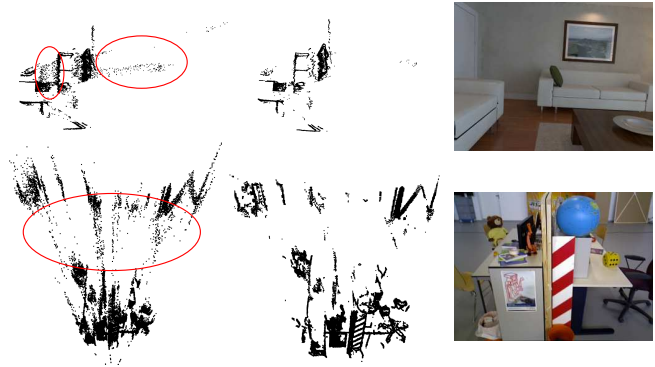} 
	\caption{Adaptive filtering on semi-dense depth map. From left to right: (back-projected) semi-dense depth map from LSD-SLAM, filtered semi-dense depth map, and keyframe image.}
	\label{fig:raw_vs_filtered_depth}
\end{figure}

\subsection{Adaptive filtering results}

We notice that the semi-dense structure from LSD-SLAM contains spurious map points, which may worsen the dense reconstruction performance.
Figure~\ref{fig:raw_vs_filtered_depth} shows a qualitative comparison between the semi-dense depth maps by LSD-SLAM and the filtered depth maps, demonstrating the effectiveness of the adaptive filter in eliminating noisy depth pixels while preserving the structure of the scene.
Quantitatively, the second row and the third row of Table~\ref{table:error_terms_ablation} (labelled "f") shows about $ 5 \% $ improvement on using adaptive filtering in dense reconstruction (see also the last four rows).

\subsection{Cost function analysis} \label{subsec:cost_function_analysis}

Table~\ref{table:error_terms_ablation} shows the reconstruction results using different combinations of error terms in the cost function.
To ensure consistent measurement of the reconstruction accuracy using different cost functions, the keyframes---i.e., the semi-dense depth and depth variance maps, and the camera poses---are pre-saved so that the densification process is not influenced by the inconsistency between runs from LSD-SLAM.
Consistent with the finding in DeepFusion, incorporation of CNN depth gradient consistency and CNN depth consistency improves the reconstruction accuracy dramatically, although our CNN does not explicitly predict depth gradient and depth gradient variance maps (see the second and last row).
However, removing the CNN depth consistency term (the the third and fourth last row), in our case, leads to better reconstruction accuracy (see also the third last and last column of Table~\ref{table:reconstruction_results});
the added generalized Charbonnier function (the second row, and labelled "c") also increases the accuracy.

\begin{table}
	\begin{center}
		\caption{Comparison of depth prediction CNNs accuracy being used in CNN-SLAM (Laina~\cite{laina}) and our system (VNLNet~\cite{vnlnet} and MiDaS~\cite{midas}) on the ICL-NUIM dataset~\cite{iclnuimdataset} and the TUM RGB-D dataset~\cite{tumrgbddataset}. (TUM/seq1: fr3\_long\_office\_household, TUM/seq2: fr3\_nostructure\_texture\_near\_withloop, TUM/seq3: fr3\_structure\_texture\_far, abs: absolute depth prediction CNN, rel: relative depth prediction CNN.)  }
		\label{table:depth_predictions_accuracy}
		\begin{tabular}{c | c c c} 
			\Xhline{2\arrayrulewidth}
			& \multicolumn{3}{c }{Percentage of correct depth (\%)} \\
			\cline{2-4}
			Sequence         & Laina (abs)       & VNLNet (abs)*     & MiDaS (rel)*      \\ 
			\hline
			ICL/office0      & \textbf{17.194 }  & 11.791            & 13.059            \\ 
			ICL/office1      & 20.838            & \textbf{45.866}   & 42.980            \\  
			ICL/office2      & 30.639            & \textbf{55.180}   & 55.136            \\
			ICL/living0      & 15.008            & 40.294            & \textbf{54.287}   \\
			ICL/living1      & 11.449            & 55.806            & \textbf{72.139}   \\ 
			ICL/living2      & 33.010            & 59.367            & \textbf{67.130}   \\
			TUM/seq1         & 12.982            & 47.552            & \textbf{54.860}   \\
			TUM/seq2         & 15.412            & 33.143            & \textbf{55.136}   \\
			TUM/seq3         & 9.450             & 52.144            & \textbf{57.255}   \\ 
			\hline
			Average          & 18.452            & 44.571            & \textbf{52.442}   \\
			\Xhline{2\arrayrulewidth}
		\end{tabular}
		\begin{tablenotes} 
			\item *After scale- and shift-correction
			
		\end{tablenotes}
	\end{center}
\end{table}

\begin{figure}
	\centering
	\includegraphics[scale=0.9]{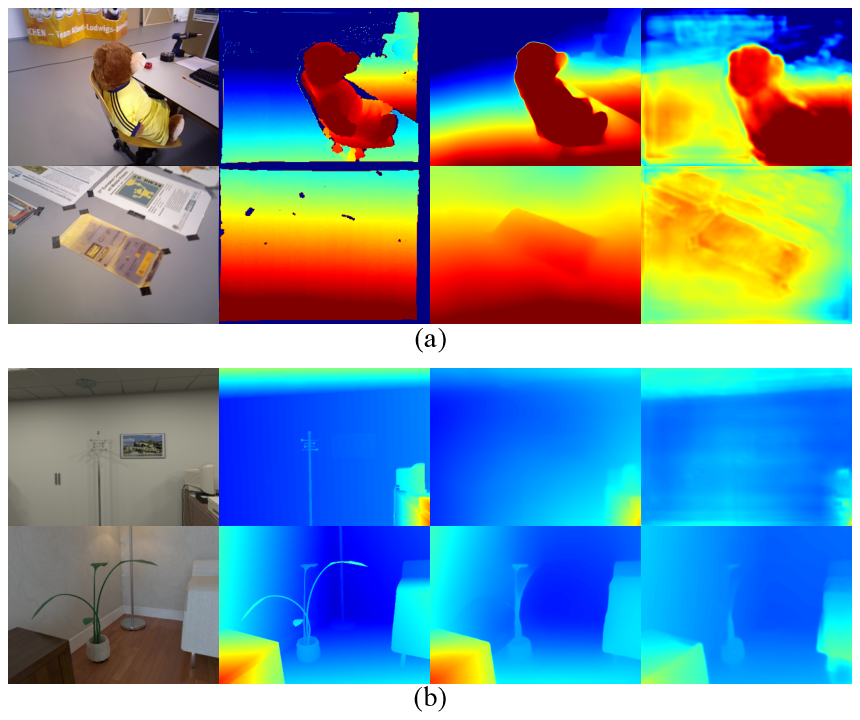} 
	\caption{Qualitative comparison of relative depth maps from MiDaS and absolute depth maps from VNLNet on (a) the TUM RGB-D dataset and (b) the ICL-NUIM dataset. From left to right: image, ground truth depth map, depth prediction from MiDaS, and depth prediction from VNLNet. }
	\label{fig:abs_ord_depth_qualitative}
\end{figure}



\subsection{Relative depth prediction vs. absolute depth prediction} \label{subsec:ord_vs_abs_depth}
To illustrate the advantage of using relative depth prediction CNNs (e.g., MiDaS), we perform the same densification step with an absolute depth prediction CNN, VNLNet\footnote{One important consideration in selecting a competing absolute depth prediction CNN is the runtime memory requirements. VNLNet is considered state-of-the-art at the time of experimental setup with a reasonable memory footprint.}~\cite{vnlnet}, and then compare the reconstruction accuracy between them. 
To promote a fair comparison, neither MiDaS nor VNLNet has been trained on the TUM RGB-D and ICL-NUIM datasets.
In Table~\ref{table:reconstruction_results}, we show that, in general, using scale- and shift-corrected relative depth prediction (labelled "MiDaS") instead of absolute depth prediction (other columns) has superior dense reconstruction performance, as a result of more accurate depth prediction from MiDaS than depth prediction from VNLNet (last and second last column of Table~\ref{table:depth_predictions_accuracy});
Laina (second column of Table~\ref{table:depth_predictions_accuracy}), another absolute depth prediction CNN being used in CNN-SLAM, is significantly less accurate than MiDaS, which indicates that the outperformance of our system may just simply be due to the fact that MiDaS provides more accurate depth prediction for densification.
Not only are the scale- and shift-corrected relative depth maps from MiDaS metrically more accurate than the absolute depth maps from VNLNet, but the relative depth maps also appear to be smoother (see Figure~\ref{fig:abs_ord_depth_qualitative}).



\subsection{Keyframe trajectory accuracy}
Table~\ref{table:ate} shows the camera tracking accuracy between our method and CNN-SLAM\footnote{Only CNN-SLAM has the ATEs on the evaluation datasets.}.
From the first two columns, we can see that our camera tracking performance, even without the pose-graph refinement, reduces the ATE of CNN-SLAM by almost $50\%$.
Since both of the systems are built upon LSD-SLAM, the performance difference could be due to our configuration settings in LSD-SLAM (see Section~\ref{sec:implementation}).
To evaluate the effectiveness of pose-graph refinement, the last column shows a baseline performance of refining the pose-graph using ground truth depth.
In general, pose-graph refinement reduces the ATE significantly to the extent that, in certain sequences, it is similar to that obtained by pose-graph refinement using the ground truth depth.

\begin{table}
	\begin{center}
		\caption{Comparison of absolute trajectory error on the ICL-NUIM dataset~\cite{iclnuimdataset} and the TUM RGB-D dataset~\cite{tumrgbddataset}. (TUM/seq1: fr3\_long\_office\_household, TUM/seq2: fr3\_nostructure\_texture\_near\_withloop, TUM/seq3: fr3\_structure\_texture\_far, abs: absolute depth prediction CNN, rel: relative depth prediction CNN, $^\diamond$: before pose-graph refinement, $^\circ$: after pose-graph refinement, *: (baseline) after pose-graph refinement with ground truth depth .)  }
		\label{table:ate}
		\begin{tabular}{c | c c c | c} 
			\Xhline{2\arrayrulewidth}
			& \multicolumn{3}{c }{Absolute trajectory error (m)} \\
			\cline{2-5}
			Sequence         & CNN-SLAM & Ours$^\diamond$  & Ours$^\circ$      & Ours*   \\ 
			\hline
			ICL/office0      & \textbf{0.266}    & 0.352            & 0.295            & 0.260\\ 
			ICL/office1      & 0.157    & 0.057            & \textbf{0.046}            & 0.045\\  
			ICL/office2      & 0.213    & 0.159            & \textbf{0.061}            & 0.045\\
			ICL/living0      & 0.196    & 0.057            & \textbf{0.039}            & 0.036  \\
			ICL/living1      & 0.059    & \textbf{0.017}            & 0.018            & 0.017\\ 
			ICL/living2      & 0.323    & 0.062            & \textbf{0.059}            & 0.056\\
			TUM/seq1         & 0.542    & 0.103            & \textbf{0.075}            & -   \\
			TUM/seq2         & \textbf{0.243}    & 0.261            & 0.245            & - \\
			TUM/seq3         & 0.214    & \textbf{0.108}            & 0.111            & - \\ 
			\hline
			Average          & 0.246    & 0.131            & \textbf{0.106}            & -  \\
			\Xhline{2\arrayrulewidth}
		\end{tabular}
		\begin{tablenotes} 
			\item -Not evaluated as not all the images have a corresponding depth map
			
		\end{tablenotes}
	\end{center}
\end{table}

\subsection{Timing evaluation}
On average, the CNN depth prediction and optimization require 0.15 s and 0.2 s, respectively, to complete. The measurements are taken on a laptop computer equipped with an Intel 7820HK CPU and an Nvidia GTX 1070 GPU.

\section{DISCUSSION}

This study illustrates the potential capability of combining a relative depth prediction CNN with a visual SLAM algorithm in solving the dense monocular reconstruction problem.
One of the major bottlenecks of the state-of-the-art dense SLAM systems is the accurate depth prediction requirement in the testing scene.
While the use of absolute depth prediction may help produce absolute scale reconstruction, it mostly makes sense in the context narrow application domain, such as dense scene reconstruction for self-driving cars.
With the proposed use of relative depth prediction, we improve the versatility of our system by forgoing absolute scale reconstruction, which can be easily recovered using fiducial markers or objects with known scales. 
With accurate relative depth prediction as well as continuous expansion in single-image relative depth CNN training datasets, we are getting closer to solving dense monocular SLAM \textit{in the wild}---dense scene reconstruction on arbitrary image sequences.

\section{CONCLUSION}

In this paper, we have presented a real-time dense SLAM system, named DeepRelativeFusion, that exploits the depth and depth gradient priors provided by a relative depth prediction CNN.
Our system densifies a semi-dense structure provided by LSD-SLAM through a GPU-based energy minimization framework.
Through ablation study, we have validated the effectiveness of the cost function used for densification, which examines the contribution of the error terms to the dense reconstruction accuracy.
Our proposed adaptive filtering has been shown to remove spurious depth pixels in the semi-dense depth maps while preserving the structure, and this in turn improves the reconstruction accuracy.
To further improve the dense reconstruction accuracy, we have presented a technique that uses two-view consistent optimized depth maps to refine the keyframe poses.
With the accurate relative depth prediction on diverse scene types, the use of a relative depth prediction CNN is a promising step towards dense scene reconstruction in unconstrained environments.

However, the densified structure does not benefit from the refined camera motion.
Motivated by the major progress in integrating depth, pose and uncertainty predictions into front-end camera tracking and back-end bundle adjustment to continuously optimize sparse structure and camera motion~\cite{d3vo,cnnsvo,cnndepthvo}, in the future, we will look into effective ways to continuously refine dense structure and camera motion.

\end{document}